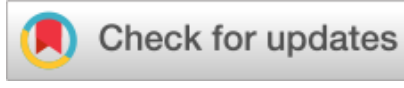

ORIGINAL ARTICLE

# Agnosticism about artificial consciousness

**Tom McClelland**

Department of History and Philosophy of Science, University of Cambridge, Cambridge, UK

**Correspondence**
Tom McClelland, Department of History and Philosophy of Science, Free School Lane, University of Cambridge, Cambridge, CB2 3RH, UK.
Email: tom.mcclelland57@gmail.com

Could an AI have conscious experiences? Answers to this question should be based not on intuition, dogma or speculation but on solid scientific evidence. However, I argue such evidence is hard to come by and that the only justifiable stance is agnosticism. The main division in the contemporary literature is between biological views that are sceptical of artificial consciousness and functional views that are sympathetic to it. I show that both camps make the same mistake of overstating what the available evidence tells us. I then consider what agnosticism means for the ethical problems surrounding the creation of artificial consciousness.

## 1 | ARTIFICIAL CONSCIOUSNESS IN CONTEXT

Could an AI have conscious experiences? Until recently, this was widely regarded as a theoretical question about an obscure science fiction scenario. Now, though, it is generally taken to be a serious question of immediate concern. The interdisciplinary literature on AI is replete with proposals regarding the prospects of artificial consciousness (AC). Government bodies are starting to take seriously the possibility that measures might be required to prevent, or at least regulate, the development of conscious AI.[1] And the public at large, fed by considerable media coverage, increasingly sees it as an urgent question. This escalation of interest is animated by the *ethical implications* of conscious AI. Schneider captures the central ethical worry:

[1]Consider the European Parliament report by Brundage et al. (2018) which includes a proposed moratorium on "synthetic phenomenology".

> Consciousness is the philosophical cornerstone of our moral systems, being central to our judgment of whether someone or something is a self or person rather than a mere automaton. And if an AI is a conscious being, forcing it to serve us would be akin to slavery (Schneider, 2019, pp. 3–4).

The question of AC has thus become not just a question that we would *like* to answer but a question we have a *moral imperative* to answer. This flurry of interest in AC takes place against the backdrop of progress in both computer science and consciousness science.

In computer science, we have witnessed numerous breakthroughs that have greatly enhanced the performance of AI. In particular, the advancement of *large language models* (LLMs) has given us AI that can sometimes appear to users to be conscious (Colombatto & Fleming, 2024). In a notable incident, Google engineer Blake Lemoine claimed that the chatbot *LaMDA* had achieved consciousness. Although very few people—experts or otherwise—agreed with Lemoine's claim, the incident contributed to a growing feeling that robust tests for AC are needed. Beyond LLMs, there have been developments with other forms of AI that raise questions about AC. Although whole-brain emulations are a long way from emulating a human brain, they can emulate the neural connectome of more simple organisms such as *Caenorhabditis elegans* and, more recently, the larval fruit fly (Bentley et al., 2016; Winding et al., 2023). If we had reason to believe these organisms are conscious, should we infer that their AI emulations are also conscious? Other programs are designed not to emulate whole brains but just the mechanisms thought to be responsible for consciousness. Would such emulations be conscious? In computer simulations of evolution, AI evolves in a manner comparable to that of organisms. Given that our consciousness emerged through an evolutionary process, might artificial evolution also give rise to conscious entities?

In consciousness science, there have been considerable developments in the scientific tools used to examine consciousness and a major expansion in the range of theories available (see Kuhn, 2024). This has contributed to a newfound confidence that questions about the distribution of consciousness can (and should) be answered scientifically. Consider the question of octopus consciousness. The science of consciousness is not in a position to *prove* whether an octopus has conscious experiences. But it is in a position to assess the *likelihood* of octopus consciousness using solid empirical findings rather than intuition, speculation or dogma (Birch et al., 2021).

So when it comes to the question of conscious AI, the dominant view is that the question can (and should) be answered scientifically. For instance, a recent major report led by Butlin et al. (2023) starts from the claim that "the assessment of consciousness in AI is scientifically tractable because consciousness can be studied scientifically and findings from this research are applicable to AI" (p. 4). We can capture this outlook with the following principle:

> **Evidentialism:** Positive or negative attributions of consciousness to AI should be based exclusively on scientific evidence.[2]

[2]What counts as scientific evidence is, of course, a contentious topic in its own right. The literature I target sets itself against: speculation without empirical grounding; reliance on intuition; and dogmatic adherence to an assumed view. In contrast, it does not preclude first-person reflection or second-person interaction from counting as evidence.

Crucially, this principle is not just meant to guide how we make judgements of AC in research contexts. It is meant to guide *real-world decisions* about AI, such as how people should treat AI, how governments should regulate AI development and even how the law should protect AI entities. Birch captures this important practical dimension of evidentialism:

> Evidence-free speculations may still have their own space elsewhere—the pub or café, or even the seminar room, book group, or lab meeting—but, for the purpose of making important, sober decisions affecting real lives, we need to create a space in which they are left at the door (Birch, 2024, p. 50).

So given that our verdicts ought to be evidence-based, what should we conclude about the prospects of AC? Different authors reach different conclusions: *AC-advocates* say that there is enough evidence to conclude that the right kind of AI would be conscious while *AC-deniers* say that there is enough evidence to conclude that AI is unlikely to be conscious. Although most authors reach their conclusions with caution, my aim is to show that even cautious conclusions are unwarranted. The evidence we have does not tell us either way whether AI could be conscious and is unlikely to do so any time soon. So, if we follow evidentialism, the only warranted stance is *agnosticism* about AC.

In the next section, I will make the case for agnosticism about AC. My argument starts from the observation that what we know about consciousness we know from human organisms. This enables us to make some warranted inferences about consciousness in non-human organisms, but when we try to extrapolate to sophisticated AI we hit an *epistemic wall*. So, although the ideal of a science-based measure of AC is a good one, the reality is that it leaves us in epistemic limbo.

In Section 3, I pit agnosticism against different approaches to the assessment of AC: The theory-heavy approach and the theory-light approach. I argue that both approaches violate evidentialism by taking a "leap of faith" that theories or markers developed with regard to organisms are also applicable to AI. I also consider the possibility that future progress in the science of consciousness might overcome this epistemic problem and argue that such optimism is unfounded. In Section 4, I address some potential objections to agnosticism: That it sets epistemic standards unnaturally high; that it is detached from the reality of how we attribute consciousness to entities; and that it relies on problematic metaphysical assumptions about consciousness. In Section 5, I move on to considering the ethical consequences of agnosticism. Not reaching a verdict on the possibility of AC leaves us with a serious dilemma about whether we should develop such AI and how to treat it if we did. I argue that the key moral difference-maker is not consciousness as such but sentience (i.e., *valenced* consciousness) and that we can get enough of an epistemic grip on artificial sentience to guide our decision-making while maintaining agnosticism.

My main aim in this paper is to make a case for agnosticism about AC. My more modest aim is to show that agnosticism is at least a serious option that deserves consideration. At present, the main division point in the literature is between *AC-advocates* who think AC is likely and *AC-deniers* who think that it is not. I seek to highlight another key division point, this time between *AC-gnostics* who believe we can reach evidence-based verdicts on AC and *AC-agnostics* who believe that we cannot. This distinctive epistemic challenge must be acknowledged as we continue to grapple with the prospect of AC.

## 2 | THE CASE FOR AGNOSTICISM

### 2.1 | AI candidates for consciousness

Before making the case for agnosticism, I should specify what kind of AI is under consideration. For three reasons, I am *not* arguing for agnosticism about current AI being conscious. First, even those favourable to AC are doubtful that current AI is conscious. For instance, Butlin et al. (2023) argue that "no current AI systems are conscious, but … there are no obvious technical barriers to building AI systems which satisfy these indicators" (p. 1). This kind of future-oriented view is common among *AC-advocates*, so an agnostic argument against them should also be targeted at future AI. Second, these doubts about current AI being conscious are well-founded. As we will see, agnosticism about consciousness in current AI is unnecessarily cautious. Third, ethical worries about AC tend to be focussed less on the AI we currently have and more on the AI we might develop (e.g., the *run-ahead principle* in Birch, 2024, p. 324). Whether we can know if future AI will be conscious is the more serious ethical question.

I will thus make a case for agnosticism about consciousness in hypothetical sophisticated AIs. My case for agnosticism is based on the problem of applying theories and markers developed in the context of organic consciousness to non-organic cases. To capture this problem, it will be helpful to focus on AIs with features that *would* constitute strong evidence of consciousness if displayed by an organism. I will call such hypothetical cases "challenger-AIs". They are challengers in the sense that they seem to be serious contenders for consciousness but also challengers in the sense that they present us with a challenging epistemic conundrum. Different approaches suggest different indicators of consciousness. Figure 1 shows the list from Butlin et al. We can stipulate that challenger-AI has all the features listed. The argument for agnosticism will show that even if we develop AI with *all* the markers of consciousness proposed by *AC-advocates*, we should *still* be agnostic.

That said, it will also be helpful to place some limits on the kind of AI under consideration. There might one day be AI that works on completely different principles from current AI and is much more akin to an organism. *AC-deniers* who regard consciousness as a biological phenomenon tend to qualify their sceptical conclusions by limiting them to *conventional* AI. Seth (forthcoming), for instance, suggests that AC might, "only be possible if we create machines that are also in some relevant sense alive" (p. 21). So, we can stipulate that challenger-AI is AI that works *on the same principles as current AI* and that displays potential markers of consciousness that *would be* considered strong indicators if displayed by an organism. I argue that if we were presented with such a challenger-AI, we should be agnostic about its consciousness.[3]

### 2.2 | The argument for agnosticism

The overall argument for agnosticism is simple:

(1) We do not have a deep explanation of consciousness.
(2) If we do not have a deep explanation of consciousness, then we cannot justify a verdict on whether challenger-AI is conscious.
(3) Therefore, we cannot justify a verdict on whether challenger-AI is conscious.

[3]I put aside here questions about *levels* of consciousness (Godfrey-Smith, 2023). This would take us into contentious territory without making much of a difference to the arguments offered.

| **Recurrent processing theory** |
| --- |
| **RPT-1:** Input modules using algorithmic recurrence |
| **RPT-2:** Input modules generating organised, integrated perceptual representations |
| **Global Workspace Theory** |
| **GWT-1:** Multiple specialised systems capable of operating in parallel (modules) |
| **GWT-2:** Limited capacity workspace, entailing a bottleneck in information flow and a selective attention mechanism |
| **GWT-3:** Global broadcast: availability of information in the workspace to all modules |
| **GWT-4:** State-dependent attention, giving rise to the capacity to use the workspace to query modules in succession to perform complex tasks |
| **Computational higher-order theories** |
| **HOT-1:** Generative, top-down or noisy perception modules |
| **HOT-2:** Metacognitive monitoring distinguishing reliable perceptual representations from noise |
| **HOT-3:** Agency guided by a general belief-formation and action selection system, and a strong disposition to update beliefs in accordance with the outputs of metacognitive monitoring |
| **HOT-4:** Sparse and smooth coding generating a 'quality space' |
| **Attention schema theory** |
| **AST-1:** A predictive model representing and enabling control over the current state |
| **Predictive processing** |
| **PP-1:** Input modules using predictive coding |
| **Agency and embodiment** |
| **AE-1:** Agency: Learning from feedback and selecting outputs so as to pursue goals, especially where this involves flexible responsiveness to competing goals |
| **AE-2:** Embodiment: Modeling output-input contingencies, including some systematic effects, and using this model in perception or control |

**FIGURE 1** Table of indicator properties from Butlin et al. (2023, p. 5) reproduced with kind permission of Patrick Butlin.

In this section, I will make a preliminary case for the two premises of the argument and unpack its conclusion. A fuller defence of the argument, including an assessment of whether the epistemic problem is temporary or permanent, will unfold in following sections.

Starting with the first premise, why think that we lack a deep explanation of consciousness? A deep explanation is one that tells us why a cognitive episode occurs consciously rather than unconsciously. Put another way, it explains why there is *something it's like* to be in a given state rather than *nothing it's like*. However, attempts to offer such an explanation run into the hard problem (Chalmers, 1995). Whenever we identify some physical or functional state associated with consciousness, it remains a mystery why that state should constitute a conscious experience rather than obtaining unconsciously. Consider a *global workspace theory* (GWT), such as the Dehaene-Changeux Model (Dehaene & Changeux, 2005), according to which a state is conscious when it is made globally available to a range of brain processes. What is it about a state being globally distributed that makes it a *conscious* state? Why are we not zombies that have a global workspace but lack subjective experience? Nothing in the theory tells us why there should be *something it's like* to be in a globally distributed cognitive state. Different theories make different claims about what kind of state suffices for consciousness, but none explains *why* it would be sufficient.

This is not to say that existing theories of consciousness have no explanatory value. Imagine we want to know why someone undergoing inattentional blindness is conscious of the basketball in their visual field but not of the gorilla.[4] A *global workspace theory* would answer (roughly) that a visual representation of the player has made it into the neuronal network that broadcasts information to other brain areas whereas a visual representation of the gorilla has not. In a shallow sense, this tells us why the subject is conscious of the player: Assuming that global distribution constitutes consciousness, the visual representation of the player is conscious because it is globally distributed. But there is a deeper question left unanswered: Why would the global distribution of this visual information constitute a visual experience? The same is true for every other theory on the table. Each theory can informatively answer various questions about consciousness but leaves the *hard* question unanswered.

Note, this invocation of the hard problem does not come with any metaphysical commitments. It is a strictly epistemic claim (see Section 4.3). Nor does it come with exceptionally high epistemic standards or a general scepticism about attributions of consciousness. It simply says that explaining phenomenal consciousness is problematic and that existing approaches do not overcome that problem. There is, of course, a considerable critical literature on the hard problem and readers will have their own view on the matter. This is not the place to litigate on that debate. My argument is that *if* you take the hard problem seriously, *then* the first premise should be plausible.

Now the second premise: If we do not have a deep explanation of consciousness, then we cannot justify a verdict on whether challenger-AI is conscious. Were we to discover a deep explanation of consciousness that solves the hard problem, we would have no trouble determining whether a challenger-AI is conscious. We would understand the necessary and sufficient conditions of consciousness and could just check whether the AI meets the conditions. But in the absence of a deep explanation, we are left only with more shallow explanations, and such shallow explanations are unsuited to the task.

Again, I will defend this claim further in due course, but for now I will illustrate the point using GWT. Studies of consciousness in humans, such as inattentional blindness studies, yield a body of evidence taken to support GWT. This theory can then give us a verdict on various difficult cases of human consciousness. Based on what we know from ordinary human cases, we can reach conclusions about whether a vegetative state patient is likely to be conscious, for example. Such conclusions would enjoy indirect empirical support from the evidence that motivated GWT in the first place. The theory gives us what you might call *inferred* conditions of consciousness: There is nothing about the theory that explains *why* global distribution would be necessary and sufficient for consciousness, but the evidence suggests that it *is* necessary and sufficient in typical human subjects and we can infer with some confidence that the same is true for atypical humans.

The evidence can even take us a step further and support conclusions about consciousness in non-human animals. If an organism has a cognitive architecture with a global neuronal workspace, then there is reason to believe it is conscious. Such attributions of consciousness would be more uncertain than the human cases. There are differences between human brains and chimp brains, and it is a live possibility that these are differences *that make a difference*. So, despite having a global workspace, chimps might lack some feature necessary for consciousness. And the further we get from humans, the greater this risk becomes. Differences between populations present us with an epistemic obstacle, and when we step over such obstacles, our confidence should decrease accordingly. But that does not mean that inferences regarding consciousness in non-human animals are unwarranted. It just means that the uncertainty of these attributions must be acknowledged.

[4]For the sake of the illustration, I put aside the possibility that the subject has a perception of the gorilla that is phenomenally conscious but not access conscious.

When it comes to AI, however, a deeper problem emerges. Even if there is good evidence that the mental states *of organisms* are conscious when broadcast in a global workspace, that is not enough to tell us that a comparable broadcasting of information *in an AI* would also be conscious. The evidence we have says nothing about whether *non-biological* global workspaces also give rise to phenomenal consciousness. When we take evidence from organic human consciousness and try to apply it to AI, our evidence hits an *epistemic wall*. The conditions of consciousness proposed by GWT are inferred, but the inferences are only warranted if kept within an appropriate scope. AI is beyond that scope.

A good way of highlighting this epistemic wall is to consider the debate between computational functionalist and biological accounts of consciousness. According to computational functionalism, consciousness is a matter of instantiating the right kind of information processes. GWT, for instance, attempts to describe the *software* of consciousness. This software happens to be running on the hardware (or *wetware*) of the brain, but if that same software were to run on silicon chips, you would still get consciousness (Dehaene et al., 2017). It is *substrate independent*. On the biological account, consciousness is a matter of having the right kind of biological state. Consciousness is an organic phenomenon that depends on how processes are physically realised. Although we might be able to simulate that phenomenon on silicon, the result would merely be a model of consciousness and not itself conscious.

Now, how can we decide between these two accounts? The putative evidence for GWT is evidence that globally distributed information is conscious *in organisms*. But when it comes to whether the same would also be true in AI, GWT hits the epistemic wall. The evidence does not distinguish between two live possibilities: (i) that the global workspace is itself sufficient for consciousness; (ii) that the global workspace is integral to how organisms achieve consciousness but *not* sufficient for consciousness by itself. If GWT offered a deep explanation of consciousness that captured *why* global distribution would constitute consciousness, then we could rule out the second possibility. But the evidence yields at best a shallow explanation that leaves both possibilities wide open. Consciousness thus presents us with a deep case of underdetermination.

The example relies on a distinction between function and substrate that has rightly come under criticism. We cannot neatly divide the mind into levels, and the functional structure of the mind is deeply intertwined with the neural structures that realise it (Cao, 2022). But the foregoing is just one of many ways to illustrate the epistemic wall. Consider Godfrey Smith's (2016) suggestion that consciousness requires a subject and that subjecthood requires an embodied organism. This presents theories like GWT with another pair of possibilities that their evidence cannot select between: (i) that a global workspace is sufficient for consciousness; (ii) that a global workspace yields consciousness only when embedded in an embodied organic subject. A range of other views, neatly summarised by Seth (forthcoming), suggest that consciousness has particular biological conditions. And each of these views can be used to set up a pair of competing hypotheses that the evidence does not allow us to choose between.

Here I have framed the problem as one of deep underdetermination: Different hypotheses yield different verdicts on AC and the evidence underdetermines our choice of hypothesis. The problem can also be framed as one of external validity: The evidence yields conclusions about causes of consciousness in the human population but extrapolations to a radically different population are unwarranted. When extrapolating we must be confident that the target population is relevantly similar to the sampled population. Potentially relevant differences should decrease the confidence with which we extrapolate. As we go from humans, to chimps, to pigs, to octopuses our confidence should be on a downward slope. But when it comes to AI the radical difference between humans and *inorganic* entities takes us over a confidence-cliff.[5]

Having made a preliminary case for the two premises, we can now move on to the agnostic conclusion. Presented with a challenger-AI, the evidence available cannot discriminate between two possibilities: (i) that the AI has these markers because it is genuinely conscious; (ii) that the AI has these markers because it mirrors conscious processes but without having subjective experience. The conclusion is not that the evidence of consciousness is mixed, with a dead heat between evidence in favour and evidence against. Rather, the conclusion is that there is *no real evidence available*. The AI has features that *would* constitute evidence were they found in an organism. But the shallow explanations available to us do not justify taking these features as evidence in non-biological cases. The only stance available to us is thus agnosticism. The term "agnostic", introduced by T. H. Huxley, seems apposite for several reasons. Huxley explains:

> I invented the word "Agnostic" to denote people who, like myself, confess themselves to be hopelessly ignorant concerning a variety of matters, about which metaphysicians and theologians, both orthodox and heterodox, dogmatise with the utmost confidence (Huxley, 1884, p. 5).

This passage reflects the spirit of my proposed view. It captures the idea that, when presented with a challenger-AI, we would be ignorant as to whether it is conscious. It also captures the idea that we should oppose *any* view that reaches a verdict with any confidence. It is not about picking between orthodox and heterodox views of AC, but rather criticising both on the grounds that their confidence is unwarranted. Huxley goes on to elaborate his outlook:

> Agnosticism is of the essence of science, whether ancient or modern. It simply means that a man shall not say he knows or believes that which he has no scientific grounds for professing to know or believe. Consequently Agnosticism puts aside not only the greater part of popular theology, but also the greater part of anti-theology. On the whole, the "bosh" of heterodoxy is more offensive to me than that of orthodoxy, because heterodoxy professes to be guided by reason and science, and orthodoxy does not (Huxley, 1884, p. 5).

Again, this reflects the spirit of my proposal. What is at issue is whether assessments of AC can be grounded in scientific evidence. The first half of this quotation captures the evidentialist principle that many in the AC literature purport to defend. The second half captures the problem that gnostics not only draw conclusions that are insufficiently supported by the evidence but do so *whilst purporting to be led by the science*. It is exactly this mismatch that my agnostic proposal is intended to highlight.

[5]Here we get into difficult questions about what kind of differences are relevant. Differences in diet between the sampled population and a target population are irrelevant so do not present an obstacle to extrapolation. But differences in brain anatomy are relevant differences that should lead us to extrapolate with less confidence. The difficulty is that determining what counts as a relevant difference is hard to do in the absence of an explanation of consciousness. This might be regarded as a problem for my claims regarding the epistemic wall. But if it is a problem, it is even *more* of a problem for the gnostic views since they rely on specific claims about which differences make a difference to consciousness. Agnosticism thus remains the more defensible view.

It is important here to distinguish *agnosticism* about AC from mere *uncertainty* about AC. Expressions of uncertainty are ubiquitous in the literature.

> [W]e can never be certain that AI is conscious, even if we could study it up close (Schneider, 2016, p. 198).

> A definitive answer to this question is not currently possible, given the lack of a consensus view about the minimally sufficient conditions for consciousness (Seth, forthcoming, p. 1).

> At this stage, we know a lot more that is relevant than we used to know. But giving a definite answer to the main questions … is not something that can be done with a lot of confidence (Godfrey-Smith, 2023, p. 7).

> We start in a position of horrible, disorienting, apparently inescapable uncertainty about other minds, and then … the uncertainty is still there at the end (Birch, 2024, p. 17).

Such expressions of uncertainty might appear quite agnostic. However, there is a world of difference between taking a *cautious* stance on whether challenger-AI is conscious and taking *no* stance. The authors above all offer a verdict one way or the other: Godfrey Smith and Seth (cautiously) deny that challenger-AI would be conscious while Schneider and Birch (cautiously) claim that it would be. But, if the argument for agnosticism is sound, conceding uncertainty in our conclusions is not enough. The only level of confidence warranted is *no* level of confidence either way. For a genuine expression of such agnosticism, we can turn to Pigliucci:

> The only examples we have of consciousness are biological. That doesn't mean that, in principle, it is not possible to build artificial consciousness, but we have no idea how to do it. And we don't know whether, in fact, it is even possible. This truly is an open question where I am entirely agnostic (Pigliucci, 2023; in Kuhn 2024, p. 147).

## 3 | AGNOSTICISM VERSUS GNOSTIC APPROACHES

The previous section offered a general argument for agnosticism about AC. This section reinforces that argument by pitting agnosticism against specific approaches to AC. The approaches available can be divided according to two cross-cutting distinctions: the theory-heavy and theory-light distinction, and the contemporary optimism versus future optimism distinction. What unites the different views is that they are all gnostic—that is, they claim that some verdict can be reached on the consciousness of challenger-AI that conforms to evidentialism.

## 3.1 | Agnosticism versus the theory-heavy approach

To adopt a theory-heavy approach to consciousness is to select a particular theory of consciousness and then assess whether a given entity is conscious according to that theory. Heavy-theorists can be *AC-advocates* or *AC-deniers*. *AC-advocates* support AC on the grounds that their preferred theory entails that a challenger-AI would be conscious. *AC-deniers* reject AC on the grounds that their preferred theory would say that a challenger-AI is not conscious. An initial problem here is that there are too many theories to choose from. Without even an approximate consensus, we do not know which theory to adopt and are left with a stalemate between contradictory proposals. The deeper problem, though, is that a theory cannot get the kind of evidential support needed to yield warranted conclusions about AC.

As discussed, any theory that gets its empirical support from human consciousness hits an epistemic wall when it tries to generalise to AC. For a theory-heavy approach to work, you need an empirically well-supported theory that yields a verdict (one way or the other) on whether a challenger-AI is conscious. But here heavy-theorists face what I call the *evidentialist dilemma*. On the one hand, they can adopt a modest version of their preferred theory that respects the epistemic wall and avoids unwarranted extrapolations to non-organic cases. But such modest theories would yield no verdict on whether a challenger-AI is conscious, leading to agnosticism. On the other hand, they can adopt a bold version of their preferred theory that specifies the necessary and sufficient conditions of consciousness as such. This theory would yield a verdict on whether a challenger-AI is conscious, but at the expense of evidentialism. It is this second horn of the evidentialist dilemma onto which most heavy-theorists fall.

We can start with how *AC-advocate* heavy-theorists compromise evidentialism. An assumption that tends to do a lot of heavy lifting (pun very much intended) is computational functionalism. Often this assumption is tacit but sometimes it is presented explicitly. For instance, Butlin et al. state: "we adopt computational functionalism, the thesis that performing computations of the right kind is necessary and sufficient for consciousness, as a working hypothesis" (2023, p. 17). Computational functionalism is the auxiliary thesis needed to get from evidence about organic consciousness to conclusions about AC. But what justifies the assumption of computational functionalism? No empirical evidence is offered in favour of it. In fact, it is not even clear what kind of empirical evidence *could* speak in favour of it. This suggests that computational functionalism is not a hypothesis arrived at by careful reflection on the evidence, but rather an article of faith. Here, the *AC-advocate* comes up against the epistemic wall, then takes *a leap of faith* over it: a leap of faith that the right computational structure suffices for consciousness. But, as Seth puts it, "the fact that computational functionalism is so frequently assumed does not mean it is true" (2024, p. 6).

Computational functionalists might respond that they do not need empirical evidence for the claim as it enjoys a priori support from broader philosophical considerations. Chalmers's famous neural replacement thought experiment offers an argument of this kind. In this scenario, the neurons in your brain are gradually replaced by silicon chips with the same functionality until eventually the whole brain is artificial. During this process, would you cease to be conscious? Chalmers and company say no and infer that computational structures suffice for consciousness. But critics respond that it simply begs the question. If computational functionalism is false, then at some point the subject *would* cease to be conscious but, due to the preservation of function, would continue to report being conscious (Udell & Schwitzgebel, 2021). This is representative of the wider philosophical debate around computational functionalism: A priori considerations cannot settle the matter any better than empirical evidence.

What about heavy-theorists who deny AC? *AC-deniers* agree with the foregoing objections to computational functionalism but make their own unwarranted claims. This is because biological theories of consciousness run into the hard problem too. Just as there is nothing about particular computational processes that explains why there is something it's like to undergo them, there is also nothing about particular *biological* processes that explains why there would be something it's like to undergo them. Consider the following passage from Godfrey Smith:

> [I]f you ask … why there should be "something it's like" to be one of these organisms to have this going on—why there should be something it feels like to have an endogenous large-scale pattern of nervous system activity, modulated by events and also put into service into the service of action from a point of view—then to me, there's not much of a gap here anymore. Once such systems exist, it should feel like something to be them (Godfrey-Smith, 2023, p. 12).

Here Godfrey Smith takes a leap of faith comparable to that of the computational functionalists. He establishes some biological features associated with consciousness and concludes that these features *suffice* for consciousness. Yet nothing about the biological processes explains *why* they should occur phenomenally. Furthermore, none of the empirical evidence warrants *inferring* that they are sufficient for consciousness. Crucially for the discussion of AC, the considerations offered also fail to warrant the conclusion that such features are *necessary* for consciousness.

The lesson here is that offering an *alternative* to computational functionalism is not the same as showing that computational functionalism is false. A biological theory-heavy approach faces the same evidentialist dilemma as its opponent. They can adopt a modest biological theory that infers how consciousness is achieved in organisms but that stays neutral on whether it might be achieved in some other way for AI. Or they can adopt a bold theory that says consciousness is *essentially* biological, yielding a clear verdict on challenger-AI but at the expense of evidentialism. The problems with computational functionalism should not push us towards *denying* AC. They should push us towards *agnosticism*. The evidence does not tell us whether to go down the functional path or the biological path, so the evidence does not tell us whether a challenger-AI would be conscious.

## 3.2 | Agnosticism versus the theory-light approach

Might a theory-light approach to consciousness fare better? This approach was developed by Griffin and Speck in the context of animal consciousness: "[A]lthough no single piece of evidence provides absolute proof of consciousness, [the] accumulation of strongly suggestive evidence increases significantly the likelihood that some animals experience at least simple conscious thoughts and feelings" (Griffin & Speck, 2004, p. 5). Without adopting any particular theory of consciousness, we can take a *meta-theoretical* approach that identifies markers of consciousness that draw on a range of theories. These markers might then be used to assess AC. This approach has some initial advantages over the theory-heavy approach. The hard problem might stop us from arriving at a single explanation of consciousness, but this approach suggests we do not need to explain consciousness to establish viable markers.

However, the fundamental problem with this approach is much the same. Why should we believe that any given marker is indicative of consciousness? Something is taken as a marker of consciousness when a theory of consciousness says that it is. And that theory should be given credence (even if not believed outright) because of the evidence in favour of it. But that evidence hits just the same epistemic wall as the theory-heavy approach. The evidence shows, at best, that these properties are markers of consciousness *in organisms*. So applying them to AI is unwarranted. This problem is captured by Shevlin:

> [T]he theory-light approach works for non-human animals insofar as it relies on the assumption that a given cluster of abilities that are consciousness-dependent in humans would be similarly consciousness-dependent if found in a non-human animal. This assumption is broadly plausible to the extent that there are broad homologies between human and animal cognitive architectures. Yet such homologies or even structural similarities are unlikely to apply when considering non-biological systems (Shevlin, forthcoming, p. 4).

So the light-theorist faces the now-familiar evidentialist dilemma. They can adopt a modest set of markers that are well-supported by the empirical evidence but only applicable in the domain where the evidence was gathered (viz., organisms). Or they can adopt a bold set of markers applicable to biological and non-biological cases but which are no longer warranted by the empirical evidence. Once more, a leap of faith is taken. This time it is not a leap of faith that a given theory is true. Instead, it is a leap of faith that the available theories *taken as a whole* somehow overcome the epistemic wall when none of them could do so individually. The wall is less salient here because the markers are arrived at on the basis of numerous theories, and each theory hits the wall in a different way. But hit it they do. So, a theory-light approach is similarly unable to reach a verdict on AC without compromising evidentialism.

## 3.3 | Agnosticism versus future optimism

So far, the approaches discussed have offered a verdict on AC based on the *current* state of consciousness science. A possible response to the agnostic arguments above is to concede that although we are in a bad epistemic position today, future research will allow us to reach empirically informed verdicts about AC. For the heavy theorist, that would mean establishing the correct theory of consciousness. For the light theorist, that would mean establishing a more robust set of markers. This gives us the set of options represented in Figure 2.

Expressions of future optimism are widespread. Seth (forthcoming), for example, suggests that "as we become better able to explain, predict, and control properties of consciousness in terms of underlying mechanisms, the sense of mystery about how consciousness relates to matter will gradually dissolve and may eventually disappear" (p. 20).

I suggest that such optimism is unwarranted. The problem for gnostic views is that efforts to understand AC face an epistemic wall. Why should we think that science is on a trajectory to overcome this wall? Consider the sense of mystery that motivated Huxley's agnosticism about the underpinnings of consciousness:

> How it is that anything so remarkable as a state of consciousness comes about as a result of irritating nervous tissue, is just as unaccountable as the appearance of the djinn when Aladdin rubbed his lamp in the story (Huxley, 1869, p. 178).

| | Contemporary optimism | Future optimism |
|---|---|---|
| **Theory-heavy approach** | We have a sufficiently well-supported theory of consciousness that can deliver verdicts on challenger-AI | We *will* have a sufficiently well-supported theory of consciousness that can deliver verdicts on challenger-AI |
| **Theory-light approach** | We have a sufficiently well-supported set of cross-theoretical markers that can deliver verdicts on challenger-AI | We *will* have a sufficiently well-supported set of cross-theoretical markers that can deliver verdicts on challenger-AI |

**FIGURE 2** Four approaches to artificial consciousness (AC).

One might argue that the ensuing 150 years of scientific progress have lessened the mystery and, if allowed to continue, will overcome it entirely. There is a sense in which we have clearly made considerable progress in the science of consciousness. But when it comes to the *hard* problem, it is not clear we have made any progress at all. We still have nothing that begins to explain why some states would give rise to phenomenal consciousness.

New theories simply *relocate* the apparent miracle of consciousness without managing to make it any less miraculous. And without progress on this matter, our efforts to understand AC will keep hitting the epistemic wall between organic evidence and artificial applications. So optimism that we will be able to reach a verdict on AC in the future is unwarranted on inductive grounds. Of course, one could just *choose* to be optimistic and not worry about whether it is warranted. But that would be its own violation of evidentialism. Once more we find gnostics taking a leap of faith—this time faith that the scientific method will overcome all obstacles in its path.

In philosophy of religion, we find a distinction between soft agnosticism, which says we *do not* know whether God exists, and hard agnosticism, which says we *cannot* know whether God exists (Moreland & Craig, 2017). Does my case for agnosticism about AC lead us to *hard* agnosticism about AC? Kuhn advocates such a view: "after super-strong AI exceeds some threshold, science could never distinguish, not even in principle, actual inner awareness from apparent inner awareness" (2024, p. 152). The issue with "in principle" problems is that they sometimes turn out to be an artefact of our ignorance (McClelland, 2020; Stoljar, 2006). Lots of phenomena seemed scientifically inexplicable *in principle* right up until they were explained. So, we should not be so confident that the problems highlighted are insurmountable. It could be that we are much closer to overcoming the hard problem than it seems and that a deep explanation of consciousness is just around the corner. But that does not mean we should be optimistic either. Faced with obstacles that we do not know how to overcome, and that we make no apparent progress in overcoming, we must take seriously the possibility that the knowledge we desire is unattainable.

The appropriate stance to take is then a *hard-ish* agnosticism that acknowledges serious obstacles to knowledge that might transpire to be surmountable but might not. The theme of this paper is humility, and it may look like the most epistemically humble view is hard agnosticism. But hard agnosticism requires us to *know* that the problems highlighted are insurmountable. Given our position of uncertainty, the most appropriate position is the hard-ish agnosticism described. It might be that a revolution in consciousness science will overcome the epistemic wall, but there is no good reason to think that such a revolution is coming.

## 4 | OBJECTIONS TO AGNOSTICISM

I now move on to consider, and rebut, three likely objections to agnosticism about AC.

### 4.1 | The untenable standards objection

The first objection is that agnosticism sets the standards of justification too high. The standards are too high if they *over-generate*, that is, lead to agnosticism in cases where agnosticism is clearly inappropriate. There are two ways in which an argument for agnosticism could over-generate: It might entail agnosticism in cases where we ought to attribute consciousness or might entail agnosticism in cases where we ought to deny consciousness.

The first kind of error might occur in the context of non-human animals. If lacking a deep explanation of consciousness entails that I cannot know that a challenger-AI is conscious, does it also entail that I cannot know that a chimp is conscious? After all, without a deep explanation of consciousness, we cannot rule out the possibility of chimps being zombies that display the biological and behavioural markers of consciousness without having subjective experience. Yet there seem to be very good empirical reasons for attributing consciousness to chimps. So if the argument for agnosticism entails that we should not attribute consciousness to chimps, we have reason to doubt the argument.

An example of the second kind of error is consciousness in LLMs. Without a deep explanation of consciousness, we do not know how consciousness might be achieved. LLMs might be very different in functional structure and physical constitution from organisms that we take to be conscious. But the evidence available cannot rule out the possibility that LLMs are conscious. Yet it seems there are good reasons to deny that LLMs are conscious. Aru et al. convincingly argue that LLMs simply “mimic signatures of consciousness that are implicitly embedded within the language that people use to describe the richness of their conscious experience” (2023, p. 1015). So, if the argument for agnosticism entails that we should not deny that LLMs are conscious, we have reason to doubt the argument.

In response to this line of objection, I maintain that my argument for agnosticism about challenger-AI does not entail agnosticism about other cases. There *are* agnostic positions that *do* entail agnosticism in other cases (Steiglitz, 2024; in Kuhn 2024, p. 56). But this is not the kind of agnosticism entailed by my argument. Without a deep explanation of consciousness, we can use what we know about human consciousness to *infer* markers of consciousness in other organisms. The problem is that when we try to extend this to non-organic cases, we hit an epistemic wall. Claims about chimp consciousness are on the right side of this wall, so my arguments are quite consistent with attributions of chimp consciousness being well supported.

My argument is also consistent with denying LLM consciousness. We cannot *prove* that LLMs lack consciousness, but that is a low bar. After all, the difficulties with disproving consciousness are what allow panpsychists to speculate that *everything* is conscious. The point is that the putative evidence in favour of LLM consciousness can be debunked. For example, linguistic statements that *would* be indicative of consciousness in humans prompt us to consider whether they are manifestations of consciousness in LLMs. But once we understand the process responsible for those reports—a process completely unlike that responsible for human reports of consciousness—we know not to trust the marker.[6] Of course, you *could* still maintain that the LLM is conscious, but such a claim would be speculative rather than evidence-based. Denying LLM consciousness is thus appropriate. This is quite different from the epistemic situation with challenger-AI. When such an AI makes linguistic reports indicative of consciousness, we cannot *debunk* the evidence as such. After all, the process underlying those reports has salient similarities to the process in humans and salient differences. But thanks to the epistemic wall, we do not know whether those differences *make a difference*. So in difficult cases like this, the appropriate response is agnosticism.

Another way to put the objection is that, when making inferences about consciousness, epistemic walls are everywhere. You can say that cautious inferences over those walls are permissible, or you can say that any such wall entails agnosticism. But what you cannot do is arbitrarily give the wall between the biological and artificial a special status. Against this objection, my aim has been to show that this special treatment is far from arbitrary. Our consistent policy should be that greater differences between the sampled population and target population should lead to lower confidence in assessments of consciousness. In other words, the confidence of our inference should be inversely proportional to the size of the epistemic wall. The wall between the biological and the artificial is huge, so if we apply that policy, we should have virtually no confidence in the resulting consciousness assessments. The key lesson here is that the case for agnosticism does not rely on onerous epistemic standards. Our ordinary standards of attributing consciousness—the standards we use when making warranted judgements about consciousness in other humans and animals—cannot be met when assessing the consciousness of challenger-AI.

## 4.2 | The ivory tower objection

The next objection is that any reservations we might have about challenger-AI being conscious will disappear when actually faced with such an AI. After all, the defining feature of a challenger-AI is that it will show all outward signs of being conscious. It will *claim* to be conscious in just the way humans do (Graziano, 2014; in Kuhn, 2024, p. 150). Even with today's more rudimentary AI, we are already seeing people attribute consciousness to programs like *Replika* and robots like *Sophia* (Shevlin, forthcoming). So it is very likely that more advanced AI will invite even stronger attributions of consciousness. A philosopher might sit in their ivory tower and argue for agnosticism, but their epistemic worries would be detached from the reality of how we actually attribute consciousness.

There are two ways of reading this objection. On the first reading, it is a claim about *evidence*. The idea is that when we actually interact with such an AI, we will have strong evidence

[6] If an LLM were shown to have a cognitive architecture like ours—such as showing signs of a global workspace—then denying consciousness would not be so straightforward. Such an LLM would be closer to the kind of challenger-AI that I target, so an agnostic view could be warranted.

of its consciousness. Tye advocates such a view in the context of the fictional android Commander Data:

> [W]e have evidence that he feels anger, fear, and pain; that he has various perceptual experiences; that he feels moods such as depression and happiness. So we have (defeasible) reason at least to prefer the hypothesis that Data has the experiences and feelings he appears to have to the hypothesis that he does not (Tye, 2016, p. 176).[7]

Against Tye, I suggest we have reason *not* to treat such appearances as good evidence. Something appearing conscious yields only *defeasible* evidence of consciousness (as Tye acknowledges). If we follow the argument for agnosticism through, then we should regard *being an AI* as a defeater. Data acts in ways that would be indicative of consciousness *in organisms*. But we are not justified in assuming that these same outward signs are indicative of consciousness in an inorganic entity. It is worth emphasising that this response to Tye is not based on a *general* scepticism about treating appearances of consciousness as evidence of consciousness. When a human, or indeed a non-human animal, appears to be conscious we have defeasible evidence of its consciousness. The difference is that in such cases this defeasible evidence is not usually *defeated*. This echoes my response to the untenable standards objection: The epistemic standards that underwrite the argument for agnosticism about AC are consistent with confident attributions of consciousness in many organic cases.

Another way of reading the objection is not as a claim about evidence but as a claim about human psychology. It is not about whether we would be *justified* in attributing consciousness to Data but about how we inevitably *will* attribute consciousness to Data (Shevlin, forthcoming). If that is the case, then agnosticism will find itself out of step with the general population. I think this psychological prediction could well be true. I do not, however, regard it as an objection to agnosticism. Sometimes good arguments have counterintuitive conclusions. And the lesson from the arguments offered is that if we ever come face-to-face with challenger-AI, our intuitions should not be trusted.

### 4.3 | The implausible metaphysics objection

Another likely objection is that agnosticism makes dubious metaphysical assumptions. In the context of general agnosticism about the scientific study of consciousness, Birch suggests that agnosticism, “often flows from a certain type of background philosophical picture. That picture is epiphenomenalism, according to which conscious experiences have no physical effects, and so leave no measurable imprint on the physical world. Conscious experience literally does nothing, so we cannot study it scientifically” (Birch, 2024, p. 51).

But epiphenomenalism is implausible for many reasons. Birch rehearses William James's influential argument that consciousness is an evolved trait and that it is unlikely we would have evolved a complex trait that has no causal power and so confers no evolutionary advantage. If agnosticism rests on epiphenomenalism, then any objection that undermines epiphenomenalism will also undermine agnosticism.

[7]Tye himself advocates a theory-free approach to attributing consciousness, but this is not an option for the views under discussion that require a *scientific* verdict on artificial consciousness.

In response to this objection, I concede that the epiphenomenalist route to agnosticism is a bad one. This is a route that some, including Huxley himself, have taken. But it is not the route taken in this paper. The *epistemic* claims in my argument are consistent with a wide variety of metaphysical views. For instance, it could well be that conscious states are *nothing more than* certain physical states. Those physical states are causally efficacious; therefore, conscious states are causally efficacious. The problem is with knowing *which* physical states are conscious and, more specifically, knowing whether states of an AI could be conscious. The case offered for agnosticism about AC avoids any problematic metaphysical commitments.

## 5 | THE ETHICAL IMPLICATIONS OF AGNOSTICISM

My final task is to establish the ethical implications of agnosticism. There are two incentives for determining these implications. The first is that they *matter*. The possibility of conscious AI presents us with a unique moral quandary, and it is important to work out what agnosticism means for our navigation of that quandary. The second is that doing so allows us to confront another significant objection to agnosticism. Deciding how to regulate the development of more advanced AI and how to treat it when it arrives seems to require us to make a judgement about how likely such AI is to be conscious. Agnosticism tells us we cannot make such judgements, but the practical reality may be that such agnosticism is unfeasible. As Birch puts it, "It is tempting to throw our hands aloft and say 'Maybe we'll never know!'. But practical and legal contexts force a choice" (2024, p. 11). I will call this the *forced choice objection*. My aim in this section is thus to make a positive recommendation about how to manage the possibility of AC responsibly and, in doing so, to rebut the forced choice objection. In Section 5.1, I will explain the ethical problem faced by agnosticism, and in Section 5.2, I will overcome that problem by sketching a concrete proposal for the responsible development of AI.

### 5.1 | The ethical problem for agnosticism

I started the paper by saying that consciousness is regarded as ethically salient. It would be more precise to say that a particular *kind* of consciousness is regarded as ethically salient. To be sentient is to have *valanced* consciousness, that is, conscious states that are good or bad to be in from the perspective of the conscious subject. Sentientism is the view that all (and only) sentient beings warrant our ethical consideration.[8] Beings that are unconscious, and beings that have consciousness devoid of valence, need not feature in our moral deliberations. Singer (2011) influentially argues that "the limit of sentience … is the only defensible boundary of concern for the interests of others" (p. 50). This might sound like a broadly utilitarian ethos, but we actually find similar claims in deontological ethics. Godfrey-Smith (2023) points out that in Korsgaard's ethics, sentience is what makes an entity a Kantian "end in itself", and in Nussbaum's ethics, it is a pre-requisite of "significant striving".[9]

[8]For a comprehensive exploration of the moral status of different forms of consciousness, see Shepherd (2018).

[9]There are, of course, objections to sentientism. Chalmers (2022) argues that "Vulcans"—defined as beings with consciousness but without valenced experiences—would still have moral import. Addressing this response is beyond the scope of the paper, though I will note that I do not share Chalmers's intuitions about such cases.

Sentience is thus a plausible ticket into the "moral circle". But there is a great deal of *uncertainty* about the boundaries of sentience. Knowing for sure that an octopus, say, is sentient would be an unrealistic expectation. What we can do, however, is assess how *likely* something is to be sentient and then take proportionate measures to avoid actions that would cause it to suffer were it sentient. This is the *precautionary principle* championed by Metzinger (2021), Birch (2024) and others. So when it comes to AI, we do not need to know for sure whether a given type of AI is sentient. Instead, we can determine whether there is a *risk* that it is sentient. And if there is a sufficient risk, we then take proportionate precautionary measures. This leaves open what the measures should be. It might be that when we create AI that could be conscious, we should give it the same rights as a sentient human agent. Or it might be that we should just avoid creating AI that would require such precautions to be taken. This generates an interesting debate about what measures would be proportional, but the debate only arises if we first judge there to be a significant chance of artificial sentience.

But now we get to the problem for agnosticism. My argument says that we cannot take a view either way on the chances of a challenger-AI being conscious. But if we cannot make a judgement on how likely it is to be conscious, we do not know what level of precaution would be proportionate. The agnostic acknowledgement that there is a *possibility* of consciousness is not enough. Without an assessment of how *likely* it is to be conscious, there is no way to proportion the precautions to the risk. Birch (2024) considers situations where there is "a total or near-total lack of evidence one way or the other, making it impossible to mount a credible, evidence-based case either for or against sentience" (p. 125). He thinks this is the situation we are in for nematode worms and insect larvae, but my argument suggests that this is the situation we would be in with challenger-AI.

What is to be done? One option proposed by Birch is to make such cases research priorities when "further investigation could plausibly lead to the recognition of S as a sentience candidate" (2024, p. 126). However, as we saw in Section 3, there is no reason to be optimistic that scientific research in the immediate future will enable to us to overcome the epistemic wall that blocks scientific assessments of AC. And without a scientific assessment of AC, there can be no scientific assessment of artificial sentience. It seems, then, that agnosticism leaves us in a moral limbo that is both unhelpful and unsustainable.

## 5.2 | A solution to the ethical problem

In order to solve this problem, we must reflect more on the link between consciousness and sentience. Consciousness is a necessary condition for sentience, so if we cannot acquire positive evidence of AC, then we cannot acquire positive evidence of artificial sentience. However, we could well reach conclusions about what kind of experience an AI *would* have *were* it conscious (McClelland, 2017). Chrisley suggests the following approach to the scientific study of AC: "do not attempt to explain why something is conscious, but instead explain why it is in this conscious state rather than that one" (2008, p. 132). Sidestepping the vexed question of whether a self-driving car is conscious, for example, we might be able to demonstrate that *if* it is conscious, *then* the contents of its consciousness would include perceptions of the road but not thoughts about the destination.

This opens up an interesting possibility regarding sentience. Perhaps we can reach verdicts of the following form: *If* the challenger-AI is conscious, *then* the contents of its consciousness would have positive/negative/no valence. Working out this kind of conditional conclusion would not be easy and would require further research. But that is a reason to make such research a priority.[10] Crucially, though, this would not require us to solve the hard problem. We can determine what the contents (or *likely* contents) of a cognitive state are without knowing what further features are needed to make that cognitive state phenomenally conscious.

This kind of conditional claim is enough to ground responsible decision-making. Consider a challenger-AI with computational states such that *if* those states were conscious, *then* they would have no valence. There would be no need to take precautions over the development or treatment of such AI because we can be confident of its insentience. Whilst remaining agnostic about whether it is conscious, we can make an evidence-based argument that even if it were conscious, it would not be sentient. Now consider a challenger-AI with computational states such that *if* those states were conscious, *then* they would have negative valence. Such an AI would present us with a moral dilemma. Taking precautions must be proportionate to the likelihood of sentience, but given agnosticism, we lack the evidence needed to assess that likelihood. Yet taking no precautions presents a real (if unquantifiable) possibility of doing great harm to a sentient entity. Here a different type of precaution enables us to avoid the moral dilemma. We can just avoid creating AI that would be sentient were it conscious! Schwitzgebel and Garza (2020) argue that if an AI is such that we do not know whether it is conscious, then we should avoid the moral dilemma it would present by not developing that AI. I propose a subtly different version of this principle: If an AI is such that we do not know whether it is *sentient*, we should avoid the moral dilemma it would present by not developing that AI. Crucially, this escapes the moral dilemma without placing undue restrictions on the development of useful AI. If the advancement of AI requires us to create AI that might be *conscious*, then so be it. We only have to ensure that its consciousness would not be valenced. This constraint would only be problematic if progress in AI somehow *required* us to engineer AI with states that would be valenced were it conscious. But in the absence of good reasons to think that this requirement obtains, the dilemma can be ignored.

Far from leaving us in moral limbo, agnosticism is consistent with us reaching evidence-based decisions about ethical problems. And against the forced choice objection, it can do so whilst staying neutral on whether challenger-AI is conscious.

## 6 | CONCLUSION

Without a deep explanation of consciousness, efforts to assess the likelihood of AC hit an epistemic wall. The dominant approaches to AC, whether they be favourable to AC or sceptical of it, leap over this epistemic wall and thereby compromise the evidentialist principle that they purport to defend. Widespread claims to offer science-based tests for AC are thus overstated. Although further progress in the science of consciousness might one day overcome this wall, there is little reason to be optimistic that it will do so soon. The resulting agnosticism makes ethical decisions about AI hard to reach. However, thanks to the distinction between consciousness and sentience, it should be possible to make evidence-based attributions of insentience without taking a stand on whether the AI is conscious. If the right precautions are taken, this is enough to overcome the ethical worries that animate the debate. My primary aim is to have made a strong case for agnosticism about AC. My secondary aim is to have at least shown that agnosticism is an important option that warrants further consideration and that alters the landscape of this important debate.

[10]Butlin et al. (2023) also recommend further research into valence. However, because of their assumption of computational functionalism they suggest we should be developing *computational theories* of valence. I recommend a more balanced approach that acknowledges the role that biology might play in valence.

## ACKNOWLEDGEMENTS

Thanks to audiences at the Cambridge Neuroscience Seminar, Digital Minds Reading Group, Joint Sessions and ZJU Summer Lecture Series, along with Tim Bayne, Marta Halina, Darryl Mathieson, Eric Schwitzgebel and two judicious referees for their invaluable observations.

## DATA AVAILABILITY STATEMENT

There is no data available.

## ORCID

*Tom McClelland* https://orcid.org/0000-0001-5956-1425